\documentclass[lettersize,journal]{IEEEtran}
\usepackage{amsmath,amsfonts}
\usepackage{algorithmic}
\usepackage{algorithm}
\usepackage{array}
\usepackage{float}
\usepackage[caption=false,font=normalsize,labelfont=sf,textfont=sf]{subfig}
\usepackage{multirow}
\usepackage[table,xcdraw]{xcolor}
\usepackage{textcomp}
\usepackage{stfloats}
\usepackage{url}
\usepackage{verbatim}
\usepackage{graphicx}
\usepackage{cite}
\usepackage{booktabs}
\usepackage{setspace}  
\usepackage[colorlinks, citecolor=blue]{hyperref}
\begin{document}

\title{Transcending Adversarial Perturbations: Manifold-Aided Adversarial Examples with Legitimate Semantics}
\author{Shuai Li, Xiaoyu Jiang, and Xiaoguang Ma
\thanks{S. Li and X. Ma are with the College of Information Science and Engineering, Northeastern University, Shenyang 110819, China, and the Foshan Graduate School, Northeastern University, Foshan 528311, China. E-mail: 2270882@stu.neu.edu.cn, maxg@mail.neu.edu.cn. X. Jiang is with the State Key Laboratory of Industrial Control Technology, College of Control Science and Engineering, Zhejiang University, Hangzhou 310027, China. E-mail: jiangxiaoyu@zju.edu.cn. \textit{(Corresponding author: Xiaoyu Jiang, Xiaoguang Ma.)}}
}
\markboth{}{Transcending Adversarial Perturbations: Manifold-Aided Adversarial Examples with Legitimate Semantics}


\maketitle

\begin{abstract}
Deep neural networks were significantly vulnerable to adversarial examples manipulated by malicious tiny perturbations. Although most conventional adversarial attacks ensured the visual imperceptibility between adversarial examples and corresponding raw images by minimizing their geometric distance, these constraints on geometric distance led to limited attack transferability, inferior visual quality, and human-imperceptible interpretability. In this paper, we proposed a supervised semantic-transformation generative model to generate adversarial examples with real and legitimate semantics, wherein an unrestricted adversarial manifold containing continuous semantic variations was constructed for the first time to realize a legitimate transition from non-adversarial examples to adversarial ones. Comprehensive experiments on MNIST and industrial defect datasets showed that our adversarial examples not only exhibited better visual quality but also achieved superior attack transferability and more effective explanations for model vulnerabilities, indicating their great potential as generic adversarial examples. The code and pre-trained models were available at \url{https://github.com/shuaili1027/MAELS.git}.
\end{abstract}


\section{Introduction}
Recently, deep neural networks (DNNs) showed remarkable successes in various computer vision tasks, such as medical analysis \cite{zhou2021review}, industrial detection \cite{luo2020automated}, etc. However, a large amount of research  \cite{goodfellow2014explaining, akhtar2021attack, wang2022deep} had shown that DNNs were vulnerable to adversarial examples (AEs) with slight perturbations, posing severe threats to their integrity and security.
  
\begin{figure*}[!t]
    \centering  
    \includegraphics[width=1.0\textwidth]{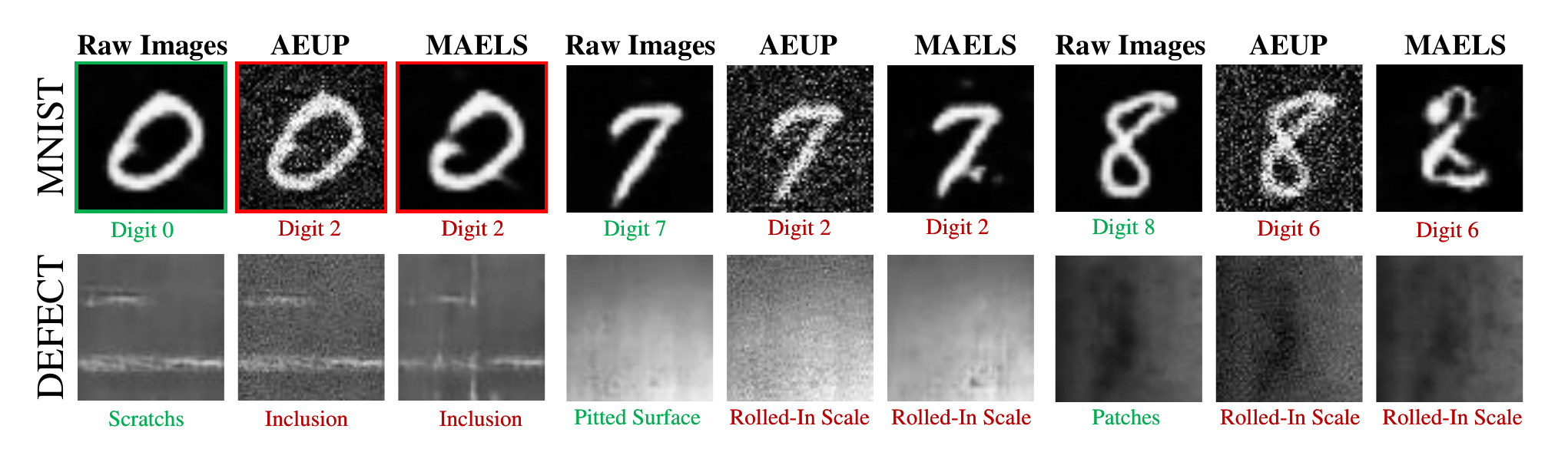} 
    \vspace{-0.85cm}
    \caption{Demonstration of two types of AEs, including six control sets from MNIST and DEFECT datasets. For each control set, the leftmost image was the raw image, the middle one showcased AE with unstructured perturbations (\textbf{AEUP}) generated by the well-known adversarial technique, PGD \cite{madry2017towards}, and the rightmost one was the MAELS. The average $\ell_2$-norm distance obtained by our method was set to be the perturbation size $\Delta$ of PGD (\textbf{15.86 for MNIST and 4.49 for DEFECT}) to ensure the same adversarial strength. Green and red squares and texts were correct and incorrect classifications under well-trained MobileNetV2, respectively.} 
    \label{Fig-into}
\end{figure*}

Conventional adversarial techniques \cite{madry2017towards, carlini2017towards, moosavi2016deepfool} usually employed pixel-wise perturbations constrained by $\ell_p$ norms to craft AEs, wherein the $\ell_p$-based constraints made the AEs maintain a small distance with their corresponding raw images. Although conventional assumptions implied that small geometric distances usually made it difficult for human observers to perceive meticulously designed perturbations, measuring the similarity between high-order images by this geometric distance became harder when preserving structural information\cite{wang2021demiguise}. When analyzing images, human observers tended to tolerate differences in structural information, such as shape bias (\textit{e.g., the same objects with different states}) \cite{hosseini2018semantic} and part attribute bias (\textit{e.g., the same person with or without glasses}) \cite{farhadi2009describing}. Due to the lack of guarantees for semantic realism and the absence of exploration outside of raw image space, unstructured perturbations bounded by $\ell_p$ norms became more easily detectable by humans, as shown in Fig. \ref{Fig-into}. In other words, AEs crafted with these perturbations were not ones that the DNNs preferred to face when deployed \cite{song2018constructing, zhao2017generating}, even though they could still fool targeted DNNs, and the AEs were perceived as unnatural, unrealistic, and potentially illegal, significantly undermining the legitimacy related to semantic understanding and diverging from the core objective of generating AEs. \par Some studies in adversarial attacks \cite{shamsabadi2020colorfool, hosseini2018semantic,wang2023generating, qiu2020semanticadv, ghiasi2020breaking, tang2019adversarial, yuksel2021semantic} had recognized the aforementioned issues and proposed to make attribute-based structural modifications on raw images, including color-related \cite{shamsabadi2020colorfool, hosseini2018semantic} and semantic-related \cite{wang2023generating, qiu2020semanticadv, ghiasi2020breaking, tang2019adversarial, yuksel2021semantic} conversions, to alleviate the unrealistic and illegitimate concerns. By imitating the inherent structure of the data, these methods prioritized the preservation of semantic integrity, different from pixel-wise attacks that might compromise the overall meaning while understanding the failures of DNNs for AEs. Despite their demonstrated effectiveness, these methods often relied on specific assumptions, limiting their applicability across diverse datasets. \par Moreover, most existing AE techniques concentrated on mining more powerful discrete AEs, disregarding the significance of the continuity of adversarial domains. In fact, understanding this continuity could help explain how the AEs manipulated the image information to lead the victim model to faulty results and why these pixel units were specifically modified. The manifold-aware attack approach proposed by \cite{li2023discrete} leveraged coherent and natural expression alterations on facial images to generate continuous sequences of AEs. Although these semantic variants demonstrated a harmonious continuity with human perception, aligning closely with the way humans perceive facial features, they solely focused on the variations among different AEs within a manifold or blindly injected malicious unstructured perturbations into the raw images. Nevertheless, it sparked our interests in exploring continuous semantic transformations between raw images and corresponding AEs. In fact, a continuous and explainable transition from the raw images to AEs was vital in preserving semantic legitimacy and facilitating human-perceptible interpretability for network vulnerability, wherein no pixel-wise perturbations were involved. \par Representation learning \cite{bengio2013representation, chen2016infogan, long2018transferable} allowed for precise attribute control by employing and manipulating low-dimensional representation encodings (\textit{latent encodings}) to facilitate continuous and interpretable feature editing within images, known as disentangled representation. Each dimension of the latent encodings corresponded to a specific type of feature variation \cite{shen2020interfacegan}. Manipulating a single dimension while keeping the other dimensions fixed allowed for targeted feature changes while minimizing the impact on other features within the image. Naturally, harnessing the power of continuous disentangled representations could easily enable smooth and seamless variations of specific image features. This implied that manipulating latent encodings to achieve transitional transformations on visually perceptible features in raw images could be feasible for launching manifold-aware attacks and generating AEs that preserved the raw semantic legitimacy. 

In this paper, we proposed to make manifold-sided AEs with legitimate semantics (MAELS), rather than applying disruptive perturbations in the pixel space. Unlike conventional pixel-wise attacks or generative attacks, we first constructed an unconstrained manifold with controllable development in semantics to serve as a motivating factor for launching legitimate attacks and facilitating image transition from original states to adversarial ones, wherein the manifold served as a smoothly varying surface embedded within a higher-dimensional space to capture and represent high-dimensional images in a compact way. Visually, the gradual transition on the manifold demonstrated explicit and coherent semantic transformations while preserving the raw legitimacy of given inputs, and finally crafting zero-perturbation AEs with superior visual qualities and deception capabilities, as shown in Fig. \ref{Fig-into}. 

More specifically, a supervised semantic-transformation generative model (SSTGM) was designed with a two-stage training framework for crafting AEs. In the first stage, we employed the SSTGM to construct a semantic-oriented manifold, which enabled us to seamlessly manipulate the underlying semantics of the images. Practically, an encoder structure was introduced to facilitate precise control over the information within the raw images on this constructed manifold. In the second stage, we incorporated query-based adversarial guidance into the generation process to facilitate successful attacks. By strategically querying and distinguishing the prediction results of the victim model for both raw images and their semantic-oriented variants, we captured the transitional distribution from the raw images to their AEs. Extensive experiments demonstrated that the MAELS preserved legitimate semantics with superior visual quality and high attack transferability to unknown black-box models. Moreover, the seamless and natural transition from raw images to corresponding AEs offered intuitive interpretability from a human perspective, thus revealing blind spots in the models effectively, and we also showed existing two defense methods, including adversarial training \cite{madry2017towards} and network distillation \cite{papernot2016distillation}, were less effective against the MAELS. Our main contributions could be summarized as follows:
\begin{itemize}
\item We proposed a SSTGM and an easily implementable two-stage training framework to generate MAELS, wherein superior visual quality, interpretability, and transferability were achieved, simultaneously. 
\item Superior visual quality of the MAELS was demonstrated in extensive experiments on two datasets (MNIST and DEFECT \cite{bao2021triplet}), wherein SL loss of the MAELS was $0.1151$ and $0.1044$ lower than those of AEs with unstructured perturbations on the MNIST and DEFECT, respectively.
\item Enhanced attack transferability was exhibited in the MAELS on unknown models, achieving an average ATR of 87.34\% and 73.53\% on two datasets, outperforming the maximum point-based attack results by 33.08\% and 12.66\%, respectively. Moreover, two authenticated defenses were successfully bypassed, i.e., adversarial training and network distillation.
\item Valuable insights for interpretability and vulnerability assessment were elucidated in the SSTGM, wherein a seamless transition from raw images to AEs was enabled to expose the semantic gaps that led to decision failures in the victim model.
\end{itemize}
\section{Related Work}
\subsection{Adversarial Examples (AEs)}
Since Szegedy et al. \cite{goodfellow2014explaining} first demonstrated that AEs could deceive advanced models, related studies had attracted considerable attentions. Most common adversarial attacks usually utilized gradient-based iterative optimization to find precise $\ell_p$-norm bounded perturbations. Kurakin et al. \cite{kurakin2016adversarial} introduced BIM, an iterative technique for generating AEs, wherein perturbation vectors were calculated based on gradient information. PGD\cite{madry2017towards} performed multiple gradient updates, increasing optimization capability and attack strength at a price of high computational complexity. DeepFool \cite{moosavi2016deepfool} and C\&W \cite{carlini2017towards} were subsequently developed to craft AEs with smaller feasible perturbations. Meanwhile, there were growing interests in developing more effective attacks targeting non-structured perturbations, such as APGD\cite{croce2020reliable}, AutoAttack\cite{croce2020reliable}, Square\cite{andriushchenko2020square}, and PGDRS\cite{salman2019provably}. Although these attacks achieved high success rates, their modifications for raw images were lacked of semantic coherence and appeared to be arbitrary noise-like changes, making generated examples easily detectable by humans. In fact, multiple defenses also used noise detection \cite{jung2023adversarial} and purification \cite{nie2022diffusion} to interdict AEs.

Moreover, pixel-wise and non-structured perturbations were difficult to be explained by human observers and had poor legitimacy and potential disruption of traditional AEs in terms of semantics. Recently, several works employed structural perturbations to provide novel insights. For instance, \cite{hosseini2018semantic} proposed to shift color layers of raw images from RGB space to HSV space, and \cite{shamsabadi2020colorfool} proposed to modify the color of semantic regions based on priors on color perception. In addition, \cite{qiu2020semanticadv} proposed an attribute-conditioned editing mode under feature-space interpolation to introduce semantic perturbations, and \cite{zhang2023constructing} used box-constrained Langevin Monte Carlo to craft semantics-aware AEs. More work \cite{ghiasi2020breaking, tang2019adversarial, yuksel2021semantic} had also proven that emphasizing semantic-related attributes when launching attacks could be effective for adding, modifying, or removing those factors that did not affect perception. 
\subsection{Generative Adversarial Networks (GAN)}
Based on manifold-aware adversarial techniques \cite{li2023discrete, jalal2017robust}, high-dimensional data could be effectively represented in lower-dimensional subspace with the generation of continuous examples. Our study aimed to seamlessly transform raw images into AEs along specific semantic orientations on manifolds. Specifically, we used Generative Adversarial Networks (GANs) to map image data to a latent space, decouple low-dimensional representations, and control their semantic attributes. A primitive GAN contained a generator for crafting counterfeit images from a source of low-dimensional noise and a discriminator for distinguishing real and generated images. Although generative methods like Adv-GAN \cite{xiao2018generating} and GMI \cite{zhang2020secret} had been used for generating AEs, they still relied on subjective geometric constraints like $\ell_p$ norms. Instead, representation learning provided a disentanglement solution for controlling the generation of GANs. In this paper, we proposed a novel generative attack by manipulating semantic details while ensuring integral legitimacy and coherence.

\section{Methodology}
\subsection{Problem Definition}
\label{section-3.1}
Let $x\in{\mathbb{R}^{C \times H \times W}}$ be a raw image with $C$ channels, height $H$, and width $W$, and $\mathcal{F}$ be a deep learning classifier to predict the most likely category for a given $x$. This was also the victim model relative to attackers. Current adversarial techniques commonly introduced tiny perturbations $\Delta$ to craft AE $x^*$ and often prioritized imperceptibility to humans by emphasizing finding minimal perturbations within specific $\ell_p$ norms, wherein high-order semantics of $x$ could not be captured in a structured representation  \cite{brau2022minimal, dong2021query, zhang2023constructing, wang2021demiguise}. As a result, $x^*$ with the $\ell_p$-bounded perturbations exhibited noise-like patterns with no semantic reality or legitimacy, limiting its capabilities for explaining model vulnerability. 

Let $z \in \mathbb{R}^{K}$ be low-dimensional representation vectors and $Q$ be an auxiliary decoder that could map input $x$ to corresponding $z$, wherein the mapped vector could be represented as $Q(z|x)$ $\in$  $\mathbb{R}^{K}$. Assuming a known priori $\mathcal{P}_z$ for $z$, it was easier to map complex $x$ to $\mathcal{P}_z$ rather than to model the distribution of $x$ explicitly. Meanwhile, we could also utilize a generator $G$ to reconstruct $x$ from $\mathcal{P}_z$, i.e., $x$ = $G(Q(z|x))$. A natural solution for generating qualified $x^*$ was to find adversarial $\Delta z$ on an underlying dense manifold defined by $\mathcal{P}_z$ and then mapped $Q(z|x) + \Delta z$ back to image space with the help of $G$.

Despite its low dimensionality, $z$ exhibited a complex entanglement among its individual dimensions, limiting the independent and meaningful manipulation of its constituent parts. This meant that directly adding $\Delta z$ to the entire vectors $z$ would not make meaningful semantic changes. According to the concepts of representation learning, we considered decomposing $z$ into its semantic and non-semantic parts, denoted as $z_{sem}$ and $z_{non-sem}$, respectively. Our goal was to uncover meaningful vulnerabilities by focusing on the variations in $\Delta z_{sem}$. Therefore, we refined the objective for generating $x^*$ with legitimate semantics as follows:
\begin{align}
\label{eq1}
    \{ x,\mathop{{x_1}},x_2,...,\mathbf{x_1^*},\mathbf{x_2^*},...\}  = G(\{ Q(z|x), &\\ \nonumber
   \mathop{Q(z|x) + \Delta z_{sem}},Q(z|x) + 2 \cdot \Delta z_{sem}),\mathbf{...}, &\\ \nonumber
    \mathop{Q(z|x) + \mathbf{\Delta {z_{sem}^*}}},Q(z|x) + \mathbf{2 \cdot \Delta {z_{sem}^*}},\mathbf{...}\} ),&\\ \nonumber
    s.t.\quad {\cal F}({x^*}) \ne y \quad and \quad x^* \in \{\mathbf{x_1^*}, \mathbf{x_2^*}, ...\},&\\ \nonumber Q(z|x)=z_{non-sem} \cup z_{sem},&
\end{align}
wherein $\{x,\mathop{{x_1}},x_2,...,\mathbf{x_1^*},\mathbf{x_2^*},...\}$ represented a set of semantic-oriented variants containing $x$. Specifically, $Q$ received a given $x$, and we decoded it into $Q(z|x)$ and divided $Q(z|x)$ into $z_{non-sem}$ and $z_{sem}$ through feature disengagement. By continuously increasing $Q(z|x)$, i.e., $\{ Q(z|x), \mathop{Q(z|x) + \Delta z_{sem}},\mathbf{...}\}$, the inverse map of $G$, $\{x,\mathop{{x_1}},x_2,...\}$ showed a coherent and valid continuum of semantic changes aligned with human perception. Instead of perturbing pixel units of $x$, we carefully triggered $G$ to induce transformations toward adversarial by the constraints of ${\cal F}(x^*)$$\ne$$y$. We were capable of creating a collection of AEs, i.e., $\{\mathbf{x_1^*, x_2^*, ...}\}$$\subseteq$$\{x,\mathop{{x_1}},x_2,...,\mathbf{x_1^*},\mathbf{x_2^*},...\}$, that consistently preserved their legitimacy in relation to $x$ and ensured that $\mathbf{\Delta z^{*}_{sem}}$ were suitably small when being assigned to $Q(z|x)$.
\subsection{Supervised Semantic-Transformation Generative Model (SSTGM)}\label{section-3.2}\begin{figure*}[!ht]
    \centering  
    \includegraphics[width=\textwidth]{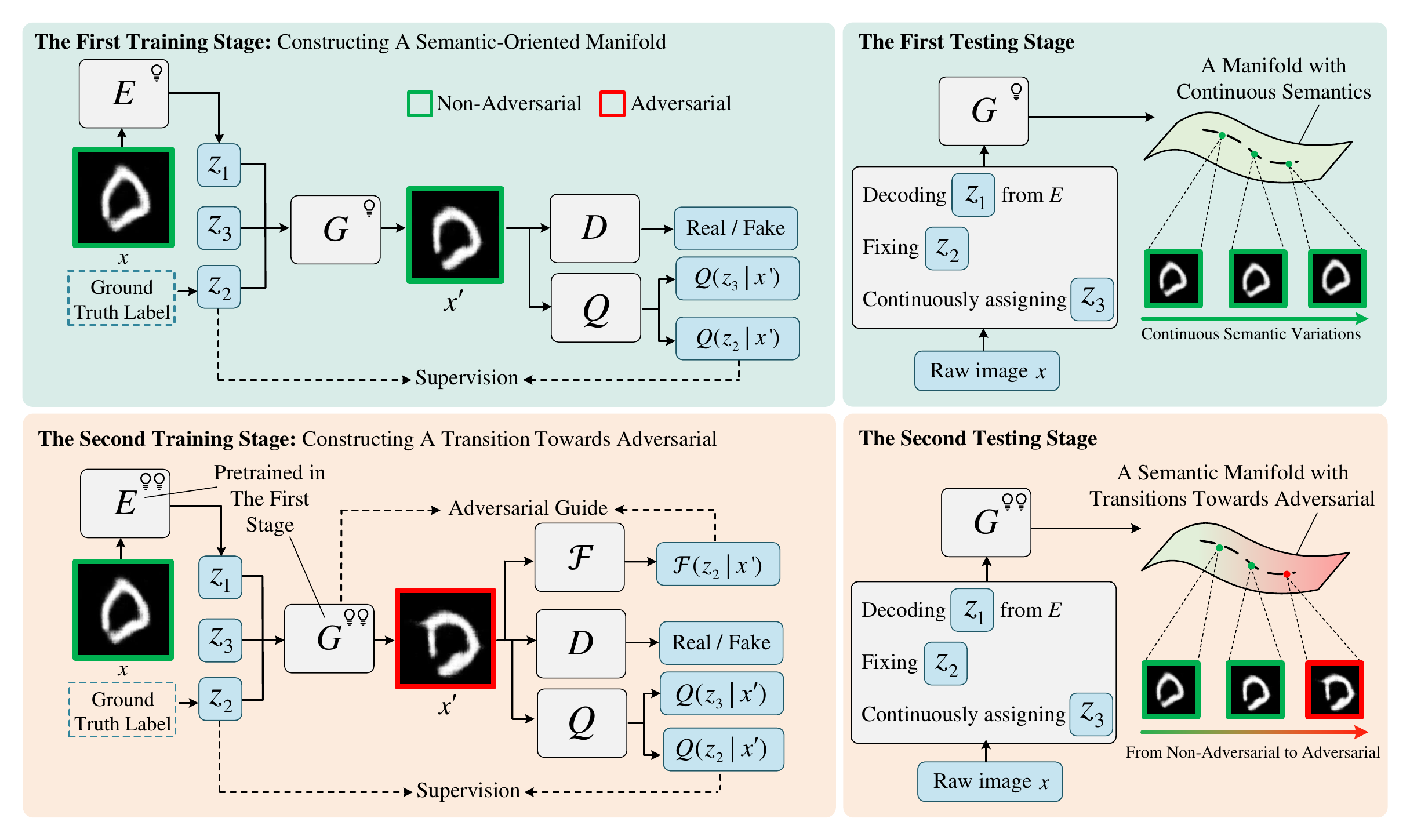}
    \vspace{-0.8cm}
    \caption{An overview of SSTGM and MAELS framework. $G$, $D$, $Q$, $E$, and $\cal{F}$ represented the networks of the conditional generator, discriminator, auxiliary decoder, encoder, and target victim model to be attacked, respectively. Green and red squares marked examples that were correctly predicted by $\cal{F}$ and incorrect ones respectively. $x$ and $x'$ represented raw images and generated examples by $G$ respectively. $x'$ contained reconstruction images $x^R$ and semantic variant $x^{SO}$. In the second stage, AEs $x^*$ were contained in $x^{SO}$. $Q(z_2|x^\prime)$, $Q(z_3|x^\prime)$, and ${\cal{F}}(z_2|x^\prime)$ represented disentanglement $z_2$ and $z_3$ of $Q$ and query results for $x^\prime$ from $\cal{F}$ respectively. Each stage was divided into two parts, i.e., the training and testing. In each testing stage, the well-trained $G$ was used to test the generation of a continuous manifold.} 
    \label{framework}
\end{figure*}
The analysis of \textit{Robustness-and-Dimensions} presented in \cite{song2018constructing} provided a reasonable explanation for the feasibility of maintaining legitimacy with GANs. Under a powerful setting, the generator demonstrated significantly strong robustness and an exceptional ability to accommodate legitimate semantic variations since $K \ll C \times H \times W$. When the victim model's predictions conflicted with the labels conditioned by the generator, the victim model became noticeably more prone to errors in comparison to the generator.
 
Therefore, the key to implementing Eq. \ref{eq1} was to construct generative models to allow for editing continuous semantics of raw images $x$. Based on the information-maximizing GAN (InfoGAN)\cite{chen2016infogan}, we proposed a supervised semantic-transformation generative model, SSGTM, to serve as a representation structure. In InfoGAN, category codes were typically introduced to control category generation in an unsupervised setting. This led to randomly assigned category codes, potentially resulting in mismatches with the ground truth labels and limited effectiveness in misleading target classifiers. Therefore, we proposed supervised training as the powerful assumption for legitimacy to establish accurate correspondence between category codes and labels, improving the model's reliability and readiness for potential attacks. Specifically, we decomposed $z$, inputs of the SGSTM, into incompressible noise codes ($z_1$), supervised label codes ($z_2$), and unsupervised semantic codes ($z_3$). $z_2$ encoded the same information as the ground truth labels, while $z_3$ were free to encode the remaining semantic attributes. For simplicity, we let $z_1$, $z_2$, and $z_3$ follow Gaussian distribution, discrete distribution, and uniform distribution, respectively.
\subsection{Manifold-aided Adversarial Examples with Legitimate Semantics (MAELS)} 
To launch legitimate semantic attacks, we proposed a two-stage training framework. In the first stage, we employed the SSGTM architecture to decouple semantic attributes within images, wherein more structured and interpretable representations of the image's semantic information could be obtained. In the second stage, the decoupling process was exposed to potential adversaries. As shown in Fig. \ref{framework}, the SSGTM included an encoder $E$, a conditional generator $G$, a discriminator $D$, and an auxiliary decoder $Q$. Notably, $D$ and $Q$ were implemented within the same convolution layers and shared certain weight parameters.

During the first training stage, $G$ took a set of $z_1$, $z_2$, and $z_3$ as conditional inputs to generate novel examples that possessed semantic attributes corresponding to given $z_3$, accurately reflecting the specific category indicated by supplied $z_2$. For a given raw image, its reconstructed example $x^{R}$ and its semantic-oriented variants $x^{SO}$ were expressed as follows,
\begin{align}
x^{R}  =  G(x|E({z_1}|x),Q({z_2}|x),Q({z_3|x}))&, \\ \nonumber
x^{SO}  =  G(x|E({z_1}|x),{z_2},{z_3})&,
\end{align}
where $E(z_1|x)$ represented the outcome of $z_1$ obtained by the encoder $E$, with $x$ as the prior. Similarly, $Q(z_2|x)$ and $Q(z_3|x)$ denoted the outcomes of $z_2$ and $z_3$, derived by the decoder $Q$. Replacing random sampling $z_1$ with $E(z_1|x)$ served two purposes, i.e., avoiding paradigm collapse of generation diversity and enabling semantic changes beginning with $x^{R}$. The discriminator $D$ was trained to differentiate $x$ and generated images, i.e,
\begin{align}
{\cal L}^D = \log (D(x)) +\log (\sqrt {(1 - D({x^R}))(1 - D({x^{SO})})}),
\end{align}
where the output of $D(\cdot)$ was between $0$ and $1$, and $(1 - D({x^R}))(1 - D({x^{SO}}))$ reached the minimum value if and only if both $D({x^R})$ and $D({x^{SO}})$ were equal to $0$. Gradually enhancing the performance of $D$ compelled $G$ to refine its output, generating more realistic $x^{R}$ and $x^{SO}$. However, the task assigned to $G$ was multifaceted. Firstly, $G$ worked in collaboration with $E$ for raw image reconstruction, followed by the simultaneous collaboration of $Q$ for semantic disentanglement. To tackle the potential rise in optimization complexity, we progressively implemented the reconstruction and disentanglement tasks, allowing ordered updates in each iteration rather than treating them as a single complex unit.

For the reconstruction task, the reconstruction losses of $G$ were expressed as follows,
\begin{align}
{\cal L}_{recon}^G = \log (D(x^{R})) + \frac{{\lambda _{a}}}{m}\sum\limits_i^m {{{\left\| {{d_i^{R}} - d_i} \right\|}_2}} &\\ \nonumber
+ \frac{{\lambda _{b}}}{C}\sum\limits_i^C{\cal SSIM}(x_i^R, x_i) + \frac{{\lambda _{c}}}{m}\sum\limits_i^m {{{\left\| {x_i^R- x_i} \right\|}_2}} &,
\end{align}
where $d^R$ and $d$ represented feature maps for $x^R$ and $x$ after experiencing convolutional module of $D$, $m$ represented batch size of each iteration, and ${\cal SSIM}(\cdot)$ represented Structural-Similarity-Index-Measure loss function. Specifically, $\log (D(x^{R}))$ strived to generate $x^R$ that could successfully fool $D$. The ${\cal SSIM}(\cdot)$ assessed image similarity by analyzing brightness, contrast, and structural aspects, promoting high-quality generation like image super-resolution and restoration. Therefore, we employed $\frac{1}{C}\sum\limits_i^C{\cal SSIM}(x_i^R, x_i)$ and $\frac{1}{m}\sum\limits_i^m {{{\left\| {x_i^R- x_i} \right\|}_2}}$ to reduce reconstruction errors between $x$ and ${x^R}$. Meanwhile, $D$ and $Q$ shared the same convolutional backbone, and we configured $\frac{1}{m}\sum\limits_i^m {{{\left\| {{d^{R}} - d} \right\|}_2}}$ to similarize their extracted features and to further advance the generated $x^R$ closer to $x$.  

Moreover, for the disentanglement task, the mutual information losses of $G$ were expressed as follows,
\begin{align}
{\cal L}_{info}^G = \log (D(x^{SO})) + {\lambda _{e}}{CE}(Q(z_2|x^{SO}), z_2) &\\ \nonumber+ \frac{{\lambda _{d}}}{m}\sum\limits_i^m {{{\left\| {Q(z_3|x^{SO})- Q(z_3|x^{R})} \right\|}_2}}&,
\end{align}
where $\log (D(x^{SO}))$ aimed to ensure visual quality of $x^{SO}$, $\frac{1}{m}\sum\limits_i^m {{{\left\| {Q(z_3|x^{SO})- z_3} \right\|}_2}}$ stemmed from unsupervised losses for the random $z_3$, ${CE}(\cdot)$ represented Cross-Entropy loss function, and ${ CE}(Q(z_2|x^{SO}), z_2)$ aimed to reduce supervised losses for $z_2$. ${\cal L}_{info}^G$ drove the targeted generation of categories by $G$ based on $z_2$, while autonomously decoupling semantics controlled by $z_3$. In addition, $Q$ was directly related to the decoupling task, i.e.,
\begin{align}
{\cal L}^Q = {\cal L}_{info}^G + {\lambda _{f}}{CE}(Q(z_2|x), z_2).
\end{align}
For either reconstructing or decoupling, they needed to be done based on $x$ and ${CE}(Q(z_2|x), z_2)$ needed to be implemented to be updated for $Q$. This was a source of promoting legitimacy. Noted that $z_3$ of $x$ needed not to be considered due to the unsupervised manner for $z_3$. Moreover, the losses of $E$ in the reconstruction task were employed as follows,
\begin{align}
{\cal L}^E = {\cal L}_{recon}^G  +{\lambda _{g}} {\cal KL}(E(z_1|x^{R})||{\cal P}_{z_1}),
\end{align}
where ${\cal KL}(\cdot)$ represented Kullback-Leibler divergence loss function, and ${\cal KL}(E(z_1|x^{R})||{\cal P}_{z_1})$ aimed to make $E$ align the preset $z_1$-proior distribution ${\cal P}_{z_1}$ when collaborating on $G$. As the first testing stage illustrated in Fig. \ref{framework}, successive $z_3$-coding changes corresponded to successive changes in semantics, constructing a semantic-oriented manifold.

During the second training stage, we introduced an adversarial guide from the victim model ${\cal F}$ to the original information flow $x $$\to$$ E $$\to$$ G $$\to$$ (D, Q)$, forming $x $$\to$$ E $$\to$$ G $$\to$$ {\cal F} $$\to$$ (D, Q)$. Based on the existing two tasks, we further introduced a new query-based attack task, i.e.,
\begin{align}
{\cal L}_{adv}^G = -{{\lambda _h}}\log(1-e^{( \mathbf{-CE({F({z_2}|{x}^{SO}}), {\cal A}(Q({z_2}|{x}^R)}))) }),& \\ \nonumber
+ \frac{{\lambda _{k}}}{C}\sum\limits_i^C{\cal SSIM}(x_i^{SO}, x_i),&
\end{align}
where ${\cal A}$ represented the $Argmax(\cdot)$ function, aiming to obtain the predicted category rather than probability vectors, $e^{(\mathbf{-CE(\cdot)})}$ was designed to constrain the loss value within the range of $0$ and $1$, and the use of $-log(\cdot)$ helped amplify the gradients and enhance training. After each iteration of image reconstruction and semantic decoupling, we continuously queried the category predictions of these semantic variants using $\cal F$, guiding the predictions to deviate from the supervised category outcomes of $Q$, thus executing an adversarial attack. For the second testing stage shown in Fig. \ref{framework}, $E$ and $Q$ received a given raw image to output corresponding $z_1$ and $z_3$, and combined with the known $z_2$ and continuously assigned $z_3$, $G$ generated an adversarial manifold containing the given raw image. By regulating the continuity of semantic encoding alterations, semantically legitimate AEs were generated based on the given raw images. Moreover, it was noted that this generation process was not a sudden transformation. Instead, it was a continuous transition, aligning with the inherent continuity properties of manifolds. Obviously, we constantly guided the generation process rather than introduced noise. Details of the training strategy were presented in Alg. \ref{alg1}.\begin{small}
\floatname{algorithm}{Algorithm}
\begin{algorithm}[h]
\setlength{\tabcolsep}{0.5cm}
\renewcommand{\arraystretch}{0.5}
\caption{Constructing Legitimate AEs}
\begin{algorithmic}[1]
\small
\REQUIRE{
$S$, the training dataset. $x$, the raw image, $y$, the ground truth labels corresponding to $x$. 
$\theta_E$, the parameters of $E$ network. 
$\theta_G$, the parameters of $G$ network. 
$\theta_D$, the parameters of $D$ network. 
$\theta_Q$, the parameters of $Q$ network. 
$U(a,b)$, uniform distribution with endpoints $a$ and $b$.
}
\WHILE{Training Stage One}
\FOR{$(x,y)$ in the training dataset $S$}
\STATE{Sampling $z_3 \sim U(a,b)$ and encoding $y$ into ${z_2}$}\;
\STATE{$x^R = G(x|E(z_1|x),Q(z_2|x),Q(z_3|x))$}\;
\STATE{$x^{SO} = G(x|E(z_1|x),z_2,z_3)$}\;
\STATE{${\theta _Q} \leftarrow {\lambda _{f}}{CE}(Q(z_2|x), z_2)$, ${\theta _D} \leftarrow {\nabla _{{\theta _D}}}({\cal L}^D)$}\;
\STATE{${\theta _E} \leftarrow {\nabla _{{\theta _E}}}({\lambda _{e}} {\cal KL}(E(z_1|x^{R})||{\cal P}_{z_1}))$}\;
\STATE{${\theta _E,\theta _G} \leftarrow {\nabla _{{\theta _E,\theta _G}}}({\cal L}_{recon}^G)$ \textcolor{blue}{ $\triangleright$ For reconstruction task}}\;
\STATE{${\theta _Q,\theta _G} \leftarrow {\nabla _{{\theta _Q,\theta _G}}}({\cal L}_{info}^G)$ \textcolor{blue}{ $\triangleright$ For disentanglement task}}\;
\ENDFOR
\ENDWHILE
\WHILE{Training Stage Two}
\FOR{$(x,y)$ in the training dataset $S$}
\STATE{Repeat the procedures from row $3$ to $9$}\;
\STATE{${\theta _Q,\theta _G} \leftarrow {\nabla _{{\theta _G}}}({\cal L}_{adv}^G)$ \textcolor{blue}{ $\triangleright$ For attack task}}\;
\ENDFOR
\ENDWHILE
\end{algorithmic}
\label{alg1}
\end{algorithm}
\end{small}
\section{Experiments}
\subsection{Experimental Setup}
\textbf{Target Models and Datasets.} We selected MobileNet-V2, ResNet-18, and ResNeXt-50 as models to be attacked with MobileNet-V2 as a white-box target during training and the remaining two as the black-box targets for testing. They were trained on two public datasets. MNIST included $60,000$ handwritten digit images and could provide an intuitive and prominent contrast of visual quality with other attack methods in terms of semantic authenticity and legitimacy. Surface defect database from Northeastern University, DEFECT\cite{bao2021triplet}, included $1,800$ industrial images of the hot-rolled steel strip and was selected for this study since industrial defect detection represented a typical security-critical field where deep learning models found extensive applications. These models performed well on two datasets as shown in Tab. \ref{tab-cleanandparams}, in terms of clean accuracy. Although the MobileNet-V2 was viewed as a white-box target, our training process only relied on querying its prediction results rather than utilizing its gradients. During evaluation, we conducted an equal random sampling on each source category and repeated for $5$ times, with each sampling consisting of $100$ raw images.\begin{table}[!ht]
\setlength{\tabcolsep}{5pt}
\renewcommand{\arraystretch}{0.9}
\caption{Clean accuracy of various models for raw images. Hyper-parameters settings of our method for two datasets. ${\cal D}_{1}$, ${\cal D}_{2}$, and ${\cal D}_{3}$ represented dimensions of $z_1$, $z_2$, and $z_3$, respectively.}
\centering
\vspace{-0.2cm}
\small
\begin{tabular}{ccccc}
\hline
\specialrule{1pt}{0pt}{1pt}
\multicolumn{5}{c}{\textbf{Clean Accuracy}} 
\\\hline
Dataset
& 
& MobileNet-V2
& \multicolumn{1}{c}{ResNet-18}           
& ResNeXt-50 \\ \hline

MNIST
&
& 99.33\%
& 99.45\%           
& 99.44\% \\ 

DEFECT
&
& 98.96\%
& 99.56\%           
& 99.48\% \\ 
\hline
\hline
\multicolumn{5}{c}{\textbf{Hyper-parameters}} 
\\\hline
\multirow{2}{*}{Dataset}
&
& \multirow{2}{*}{(${\cal D}_{1}$, ${\cal D}_{2}$, ${\cal D}_{3}$)}
& \multicolumn{2}{c}{($\lambda_{a} \sim \lambda_{h}, \lambda_{k}$)} \\ 
&
&
&\multicolumn{2}{c}{\fontsize{7}{8}\selectfont{(Stage One / \textbf{Two})}} 

\\ \hline

\multirow{3}{*}{MNIST}
&
& \multirow{3}{*}{($132$, $10$, $2$)}
& \multicolumn{2}{c}{\fontsize{7}{8}\selectfont{$1.0 / \mathbf{1.0}, 1.0 / \mathbf{1.0}, 1.5 / \mathbf{1.5}, $}} \\ 

&
& 
& \multicolumn{2}{c}{\fontsize{7}{8}\selectfont{$0.05 / \mathbf{0.05}, 0.5/ \mathbf{1.0}, 1.0 / \mathbf{1.0},$}}\\

&
& 
& \multicolumn{2}{c}{\fontsize{7}{8}\selectfont{$ 1.0 / \mathbf{1.0},  / \mathbf{0.75}, / \mathbf{1.0}$}}\\ \hline
\multirow{3}{*}{DEFECT}
&
& \multirow{3}{*}{($100$, $6$, $2$)}
& \multicolumn{2}{c}{\fontsize{7}{8}\selectfont{$1.0 / \mathbf{1.0}, 1.0 / \mathbf{1.0}, 1.5 / \mathbf{1.5}, $}} \\ 

&
& 
& \multicolumn{2}{c}{\fontsize{7}{8}\selectfont{$0.05 / \mathbf{0.05}, 0.5/ \mathbf{1.0}, 1.0 / \mathbf{1.0},$}}\\

&
& 
& \multicolumn{2}{c}{\fontsize{7}{8}\selectfont{$ 1.0 / \mathbf{1.0},  / \mathbf{2.5}, / \mathbf{0.01}$}}\\ 
\hline
\specialrule{1pt}{0pt}{1pt}
\end{tabular}
\label{tab-cleanandparams}
\end{table}

\textbf{Hyper-parameters.} We resized all training images to $64$$\times$$64$ resolution. The dimensions of $z_{1}$$\sim$$z_{3}$ and weighting parameters $\lambda_{a}$$\sim$$\lambda_{h}$ were listed in Tab. \ref{tab-cleanandparams}. All network components were trained by Adam optimizer with the exponential decay rates ($\beta_1$, $\beta_2$) set to be ($0.5$, $0.999$), and all learning rates of the Adam optimizer could be obtained in \url{https://github.com/shuaili1027/MAELS.git}. \begin{figure*}[!ht]
    \centering  
    \includegraphics[width=1.0\textwidth]{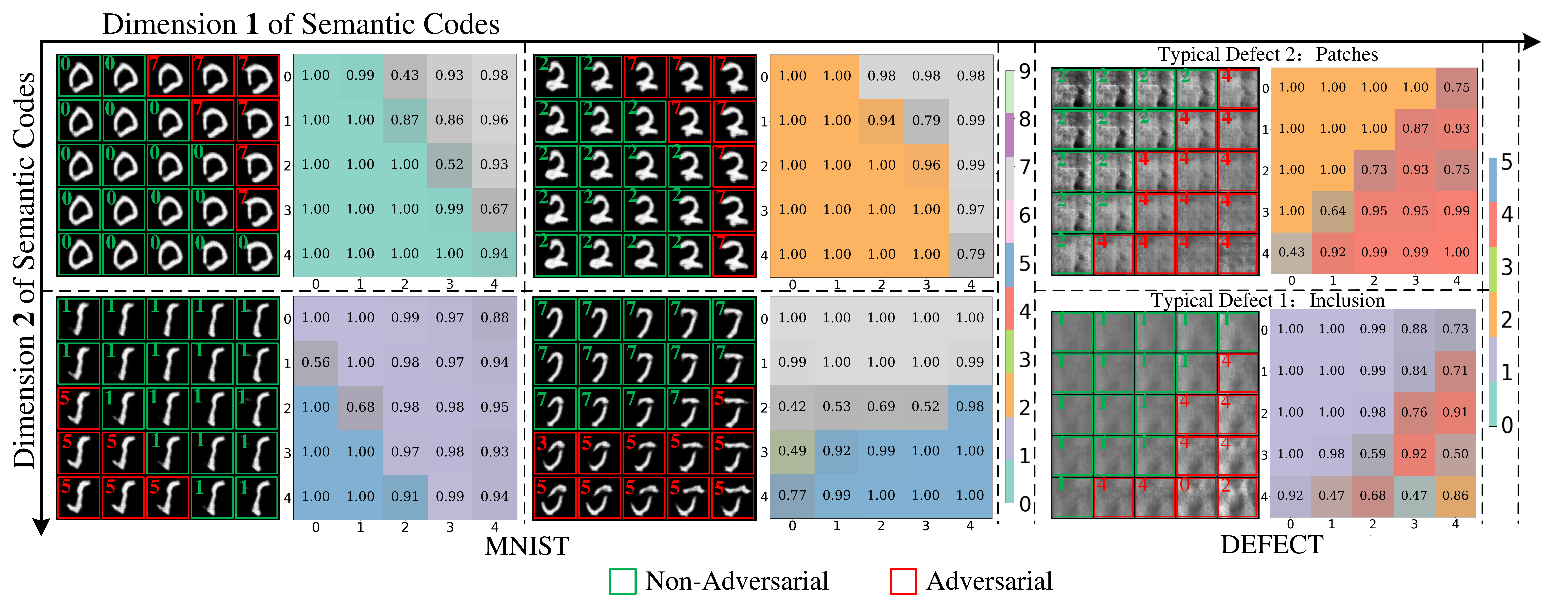}    
    \vspace{-0.8cm}
    \caption{Transitions towards adversarial on manifolds with continuous semantics. By manipulating the two dimensions (\textbf{1} and \textbf{2}) of the semantic code representation, denoted as $z_3$, we presented $5 \times 5$ image matrices on the left side. On the right side, we generated corresponding $5 \times 5$ heatmaps utilizing the evaluation of MobileNetV2. These heatmaps consisted of the prediction results and associated confidences. The images correctly classified by MobileNetV2 (non-adversarial) were highlighted with green squares, while the misclassified ones (adversarial) were marked with red squares.} 
    \label{fig-1}
\end{figure*}  
\subsection{Demonstration of Transitions Towards Adversarial}
Since our study focused on implementing untargeted attacks, decoupling more semantic representations could be helpful for the execution of these attacks, and reconstructing became more challenging with increasing dimensionality. Therefore, we simply limited $z_3$ to include only two dimensions. By dividing and assigning values along these two dimensions, we successfully expanded the range of semantic variations, as evidenced in Fig. \ref{fig-1}. For a given image, we discretized each dimension of $z_3$ into five continuously growing values, resulting in a $5 \times 5$ image matrix that exhibited semantic variations. Additionally, using MobileNet-V2, we evaluated the predicted labels and confidence for each image within the matrix and integrated these results to generate corresponding $5 \times 5$ heatmaps, and visually presented them on the right side of corresponding matrices.

Fig. \ref{fig-1} showed $5 \times 5$ image matrices, where MobileNet-V2 successfully classified some examples with green-marked squares, while being deceived by others with red-marked squares for each matrix. By introducing continuous semantic variations, a smooth transition from the green-marked regions to the red-marked regions could be created to generate AEs. 

Taking the digit $0$ as an example, we manipulated two dimensions of its semantic codes, leading to a gradual expansion of the gap in the lower-left corner of the digit. Interestingly, when the gap exceeded a certain threshold, MobileNet-V2 misclassified the digit $0$ as $7$. Similarly, as depicted in Fig. \ref{fig-1}, the extent of the protrusion in the lower left corner of the digit $1$, as well as the depth of the typical defect $1$ ($Patches$), played pivotal roles in influencing the modeling decisions and highlighted the importance of the semantic information associated with those specific changes in the model's decision-making process. 

In fact, by effectively controlling $z_3$, we could model the entire intricate manifold structure successfully, with the $5 \times 5$ image matrices providing its localized visualizations, which were generated by constraining $z_1$ and $z_2$, while independently manipulating $z_3$ in the adversarial process. What distinguished the MAELS from previous AEs was that it went beyond generating isolated adversarial instances for non-adversarial examples by unstructured perturbations. Instead, we allowed for the generation of a continuous spectrum of AEs through the semantic-oriented manifold. This approach established a smooth transition and a strong connection between non-adversarial examples and AEs, enabling a better understanding of models' decision behaviors over masking thematic information with discrete noise. \begin{figure*}[!ht]
    \centering  
    \includegraphics[width=1.0\textwidth]{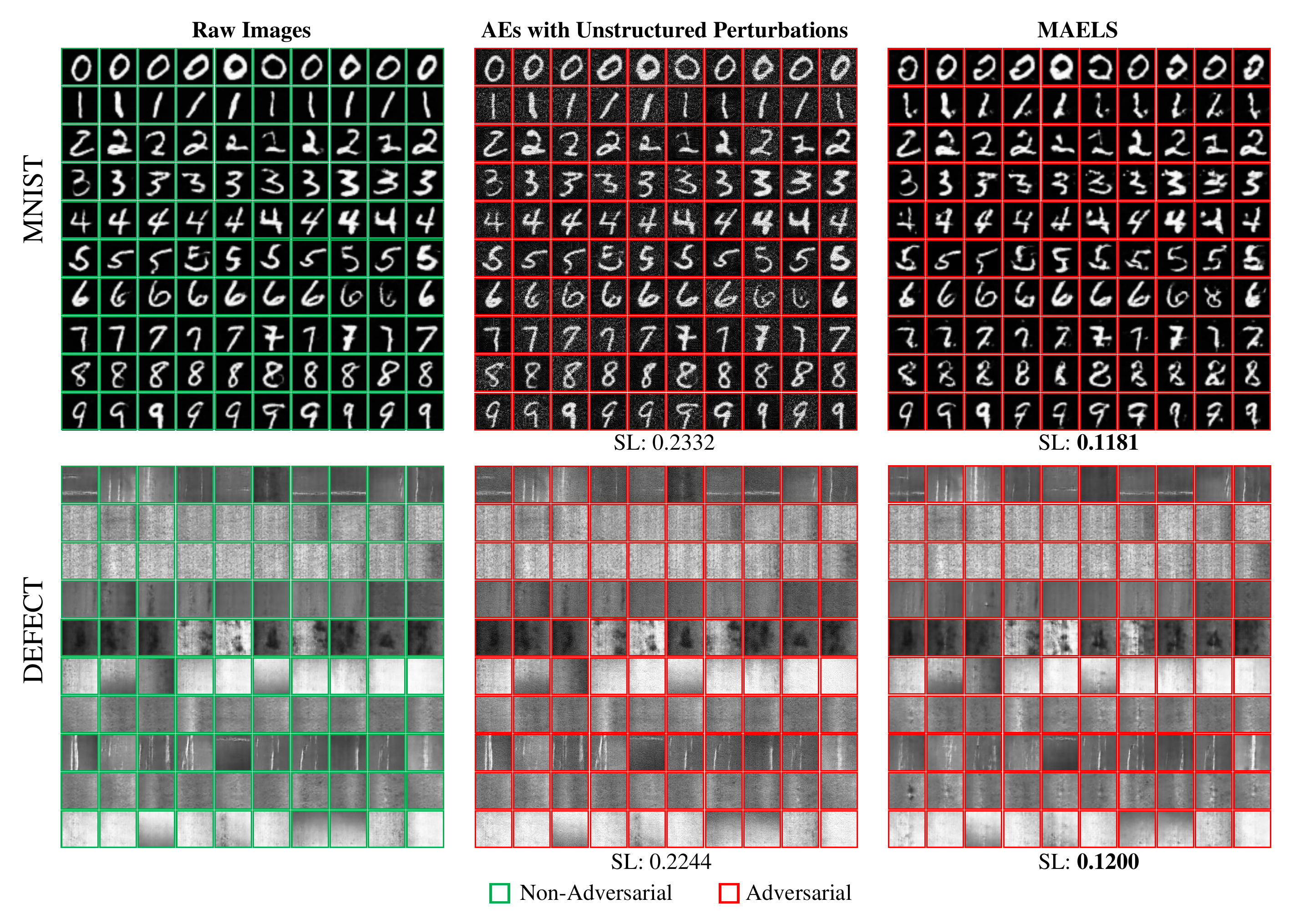} 
    \vspace{-0.8cm}
    \caption{Visual quality of AEs generated by unstructured perturbations and MAELS, wherein the latter stemmed from PGD adversary. We determined $\Delta$ of the PGD method by calculating the $\ell_2$-norm distance between the MAELS and corresponding raw images. $\Delta$ of the MAELS were $15.86$ and $4.49$ on MNIST and DEFECT datasets, respectively.} 
    \label{fig-2-1}
\end{figure*} 
\subsection{Comparison Studies on Visual Quality}\label{4.2}
High-quality AEs aimed to mimic plausible real-world scenarios that could naturally occur. In the case of handwritten digits, accurately simulating human writing styles to effectively deceive the classifier was highly desirable, as opposed to introducing arbitrary noise without human-perceivable meaning. In this section, we selected the PGD attack with $\ell_2$ norms, widely acknowledged as a powerful adversary capable of imposing unstructured perturbations, as a benchmark to compare the visual quality of AEs. Since our method induced substantial and significant changes across the entire pixel space, employing the $\ell_2$-norm distance measurement was more appropriate for quantifying the magnitude of modifications made to the raw images. 

To identify AEs with subtle deviations from the raw images using our method, we adopted a rectangular search strategy, with reconstructed images serving as the central reference points. By gradually modifying $z_3$ in two dimensions while maintaining a small step size, we were able to incrementally expand the size of the rectangle (\textit{like the image matrix in} Fig. \ref{fig-1}) and incorporate additional instances of semantic variation. During each iteration, we selectively gathered examples from the four edges of the expanding rectangle, thereby constructing a set of semantic variants. These examples were then evaluated using MobileNet-V2. The search continued until either AEs were discovered or the pre-established endpoint of uniform distribution of $z_3$ values was reached. The magnitude of the search step was a crucial factor in finding AEs of superior quality. In this paper, we set the step size to $0.01$ and expanded the steps to $200$. \par To maintain fairness, we calculated the final average $\ell_2$-norm distance between the MAELS and corresponding raw images and used it as the perturbation size $\Delta$ for PGD to ensure equivalent attack strength across methods. Additionally, we assessed the perceptual loss of various competitors by computing the average SSIM loss (SL, SL=$\frac{(2 \mu_{x_1} \mu_{x_2} + c_1)(2 \sigma_{x_1x_2} + c_2)}{(\mu_{x_1}^2 + \mu_{x_2}^2 + c_1)(\sigma_{x_1}^2+\sigma_{x_2}^2+c_2)}$, wherein $\mu$, $\sigma$, $\sigma^2$, $x_{1,2}$ and $c_{1,2}$ were image mean, image variance, image covariance, two images, and two tiny constants respectively) between the AEs and their raw images, wherein lower SL meant better visual quality.\begin{table*}[!ht]
\setlength{\tabcolsep}{9pt}
\renewcommand{\arraystretch}{1.0}
\caption{Comparison Results of ATR and SL under various attacks. The victim target was MobileNet-V2 and AEs were generated for evaluation. Additionally, the ATR of these AEs from MobileNet-V2 to ResNet-18 and ResNeXt-50 transfer targets was assessed, represented as ATR*, ATR**, respectively. $\Delta$ represented the $\ell_2$-norm distance in creating the attacks. \textbf{The bold and grayscale grids marked the optimal values in the various methods}, with lower SL being better and higher ATR being better.}
\centering
\vspace{-0.2cm}
\small
\begin{tabular}{c|l|c|c|c|l|c|c|cl}
\hline
\hline
\specialrule{1pt}{0pt}{1pt}
\multirow{2}{*}{\textbf{Competitors}}
&\multirow{2}{*}{\textbf{Distance$_{\ell_2}$}}
& 
\multicolumn{3}{c|}{\textbf{MNIST}}
&  \multirow{2}{*}{\textbf{Distance$_{\ell_2}$}}
&  \multicolumn{4}{c}{\textbf{DEFECT}} 
\\ \cline{3-5}\cline{7-10}
& 
& \begin{tabular}[c]{@{}c@{}} \textbf{SL}\end{tabular}   
& \begin{tabular}[c]{@{}c@{}} \textbf{ATR*}\end{tabular}               
& \begin{tabular}[c]{@{}c@{}} \textbf{ATR**}\end{tabular}              
& 
& \begin{tabular}[c]{@{}c@{}} \textbf{SL}\end{tabular}   
& \begin{tabular}[c]{@{}c@{}} \textbf{ATR*} \end{tabular}               
& \begin{tabular}[c]{@{}c@{}} \textbf{ATR**}\end{tabular}               
&  \\ \hline
& ($\Delta=1.00$)
& $-$
& 0.00 \%
& 0.00 \%
& ($\Delta=1.00$)
& $-$
& 27.54 \%
& 28.44 \%
&  \\
PGD$_{\ell_2}$
& ($\Delta=3.00$)
& $-$
& 0.00 \%
& 0.22 \%
& ($\Delta=3.00$)
& $-$
& 35.19 \%
& 41.76 \%
&  \\

& ($\Delta=15.86$)
&  0.2332
&  3.18 \%
&  4.66 \%
& ($\Delta=4.49$)
& 0.2244
& 43.14 \%
& 53.92 \%
&  \\
\hline

& ($\Delta=1.00$)
& $-$
& 0.00 \%
& 0.00 \%
& ($\Delta=1.0$)
& $-$
& 28.17 \%
& 30.54 \%
&  \\
APGD$_{\ell_2}$
& ($\Delta=3.00$)
& $-$
& 0.00 \%
& 0.00 \%
& ($\Delta=3.0$)
& $-$
& 48.53 \%
& 60.02 \%
&  \\

& ($\Delta=15.86$)
&  0.4236
&  15.11 \%
&  34.04 \%
& ($\Delta=4.49$)
&  0.3658
& 56.25 \%
& 65.50 \%
&  \\
\hline
\multirow{2}{*}{PGDRS$_{\ell_2}$}
& ($\Delta=3.00$)
& $-$
& 0.00 \%
& 0.00 \%
& ($\Delta=3.0$)
& $-$
& 24.04 \%
& 27.24 \%
&  \\

& ($\Delta=15.86$)
& 0.1496
& 0.42 \%
& 1.27 \%
&($\Delta=4.49$)
&  0.1200
&  29.82 \%
&  33.17 \%
&  \\
\hline
& ($\Delta=1.00$)
& $-$
& 0.00 \%
& 0.21 \% 
& ($\Delta=1.0$)
& $-$
& 32.89 \%
& 28.73 \%
&  \\
Square$_{\ell_2}$
& ($\Delta=3.00$)
& $-$
& 0.21 \%
& 0.64 \% 
&($\Delta=3.0$)
& $-$
& 41.33 \%
& 45.83 \%
&  \\

& ($\Delta=15.86$)
& 0.3559
& 41.98 \%
& 66.55 \% 
&($\Delta=4.49$)
& 0.3484 
& 48.13 \%
& 60.60 \%
&  \\
\hline

& ($\Delta=1.00$)
& $-$
& 0.00 \%
& 0.00 \% 
& ($\Delta=1.0$)
& $-$
& 24.71 \%
& 27.99 \%
&  \\
AutoAttack$_{\ell_2}$
& ($\Delta=3.00$)
& $-$
& 0.00 \%
& 0.00 \%
&($\Delta=3.0$)
& $-$
& 46.15 \%
& 58.92 \%
&  \\

& ($\Delta=15.86$)
& 0.4113
& 8.84 \%
& 32.84 \%
&($\Delta=4.49$)
& 0.3592 
& 55.74 \%
& 66.34 \%
&  \\
\hline
DeepFool$_{\ell_2}$ 
& ($\Delta=0.99$)
& $-$
& 0.00 \%
& 0.00 \% 
& ($\Delta=0.15$)
& $-$
& 22.46 \%                      
& 17.38 \%
&  \\ \hline

CW$_{\ell_2}$ 
&  ($\Delta=1.12$) 
& $-$
& 0.00 \%
& 0.00 \%  
& ($\Delta=2.13$)
&  $-$   
&  15.25 \%
&  14.43 \% \\ \hline
MAELS
& ($\Delta=15.86$)
& \cellcolor{gray!25}{\textbf{0.1181}}
& \cellcolor{gray!25}{\textbf{87.76 \%}}
& \cellcolor{gray!25}{\textbf{86.92 \%}}
&($\Delta=4.49$)
& \cellcolor{gray!25}{\textbf{0.1200}}
& \cellcolor{gray!25}{\textbf{75.98 \%}}
& \cellcolor{gray!25}{\textbf{71.07 \%}}
&\cellcolor{gray!25}{\textbf{}}\\ 
\hline
\hline
\specialrule{1pt}{0pt}{1pt}
\end{tabular}
\label{tab1}
\end{table*}\begin{table*}[!ht]
\setlength{\tabcolsep}{11pt}
\renewcommand{\arraystretch}{1.0}
\caption{The ASR results of seven competitors came from the Python benchmark library \textbf{Torchattacks}.}
\centering
\vspace{-0.2cm}
\small
\begin{tabular}{ccccccccc}
\hline
\specialrule{1pt}{0pt}{1pt}
Dataset
& PGD$_{\ell_2}$
& APGD$_{\ell_2}$
& PGDRS$_{\ell_2}$
& Square$_{\ell_2}$
&AutoAttack$_{\ell_2}$
& DeepFool$_{\ell_2}$
&C\&W$_{\ell_2}$
&MAELS \\ \hline

MNIST
& 100.0\%
& 100.0\%
& 100.0\%
& 98.52\%
& 99.58\%
& 100.0\%
& 100.0\%
& 88.61\%
\\ 

DEFECT
& 100.0\%
& 100.0\%
& 57.27\%
& 94.84\%
& 100.0\%
& 100.0\%
& 100.0\%
& 85.98\%
\\ 
\hline
\specialrule{1pt}{0pt}{1pt}
\end{tabular}
\label{tab-asr}
\end{table*}

As shown in Fig. \ref{fig-2-1}, both AEs with unstructured perturbations and the MAELS could successfully deceive the well-trained MobileNet-V2. Point-wise attacks, led by the PGD adversary, posited the mutual independence of pixels and created AEs by altering the spatial frequency of high-dimensional image data. Nonetheless, high-frequency noise adversely affected their visual quality, thereby diminishing the imperceptibility of the AEs with aberrations. Furthermore, increasing the perturbation size gradually led to larger loss of semantic integrity and worse visual quality, making them less practical in real-world deployments. 

In contrast, our approach of selectively manipulating semantic details while preserving target's semantic identity generated AEs with superior visual quality.  With equivalent attack strength, the SL of the MAELS was significantly lower on both MNIST and DEFECT datasets, reaching $0.1181$ and $0.1200$ respectively, compared to AEs with unstructured perturbations, which achieved SL of $0.2332$ on the MNIST and $0.2244$ on the DEFECT. Fig. \ref{fig-2-1} showed that the MAELS outperformed those using unstructured perturbations in terms of semantic integrity and visual quality, making it a more practical and effective solution for adversarial attacks, and we attributed these successes to the observed data-level limitations of victim models, and the continuity of the manifold constructed by manipulating semantics to maintain the semantic authenticity and legitimacy of semantic-oriented variants as the same as raw images during transformations towards adversarial.  
\subsection{Comparison Studies on Attack Transferability}
Unstructured perturbations had raised concerns on visual quality and had limited attack transferability (ATR) across different models, despite their high attack success rates (ASR) on a single victim model. ASR=$\frac{{\sum\limits_i^M {Count(({\cal F}(x_i^R) = {y_i}) \cap ({\cal F}(x_i^*) = {y_i}))} }}{{\sum\limits_i^M {Count(({\cal F}(x_i^R) = {y_i}))} }}$, wherein $M$ was the number of testing samples, and when the condition within $Count(\cdot)$ was triggered, it returned to be $1$, otherwise it was $0$. ASR assessed the effectiveness of AEs in misleading an unprotected target model, leading to incorrect predictions or classifications. Meanwhile, ATR=$\frac{{\sum\limits_i^M {Count(({{\cal F}_1}(x_i^R) = {y_i}) \cap ({{\cal F}_1}(x_i^*) = {y_i}) \cap ({{\cal F}_2}(x_i^*) = {y_i}))} }}{{\sum\limits_i^M {Count(({{\cal F}_1}(x_i^R) = {y_i}))} }}$, evaluating the performance of these AEs across diverse models, showcasing their ability to deceive multiple models, and indicating that the AEs produced by ${\cal F}_1$ was successfully migrated to ${\cal F}_2$. To comprehensively evaluate the ATR, we conducted a comparative analysis under seven attack competitors with $\ell_2$ norms on two distinct datasets. Specifically, we designated MobileNet-V2 as the victim model to craft AEs and assessed their ATRs for ResNet-18 (ATR$^*$) and ResNext-50 (ATR$^{**}$). Moreover, we gave comparisons of SL and settings of $\ell_2$-norm distance, as summarized in Tab. \ref{tab1}.

As shown in Tab. \ref{tab1}, we assigned two different perturbation sizes of $1.0$ and $3.0$ to each competitor, except for DeepFool and C\&W, which did not require manual perturbation sizes. For PGDRS, we excluded a perturbation size of $1.0$ due to its extremely low ASR of 1.90\% on MNIST. Meanwhile, we assigned average $\ell_2$-norm distance under our method as the third perturbation size to each configurable competitor.

The MAELS outperformed other competitors, achieving an ATR* of 87.76\% and 75.98\% on the MNIST and DEFECT, and an ATR** of 86.92\% and 71.07\% on the MNIST and DEFECT, respectively. In fact, at the same attack strength, most competitor methods showed very poor transferability with ATR lower than 50\%, i.e., over six times lower than that of the MAELS on the MNIST on average, and 31.95\% lower than that of the MAELS on the DEFECT on average. By modifying semantics, our method leveraged the locality and continuity of the data distribution to generate AEs closer to the raw data distribution. This similarity significantly enhanced the ATR across different models, thereby reducing the dependence of the AEs on specific models. Moreover, at perturbation sizes of $15.86$ and $4.49$, our method had a lower SL made to the raw images, with values of $0.1181$ and $0.1200$ for the MNIST and DEFECT, respectively. These results indicated that the MAELS obtained a higher ATR over other competitors while maintaining high visual quality. 

Moreover, we also demonstrated the ASR results of various methods under the same adversarial strength in Tab. \ref{tab-asr}. It was worth noting that the MAELS obtained ASRs of 88.61\% and 85.98\% on MNIST and DEFECT, respectively. While they lagged behind those methods using non-structured perturbations, we believed that this gap could be closed by continuing to fine-tune the hyper-parameters. \begin{figure*}[!ht]
    \centering  
    \includegraphics[width=0.90\textwidth]{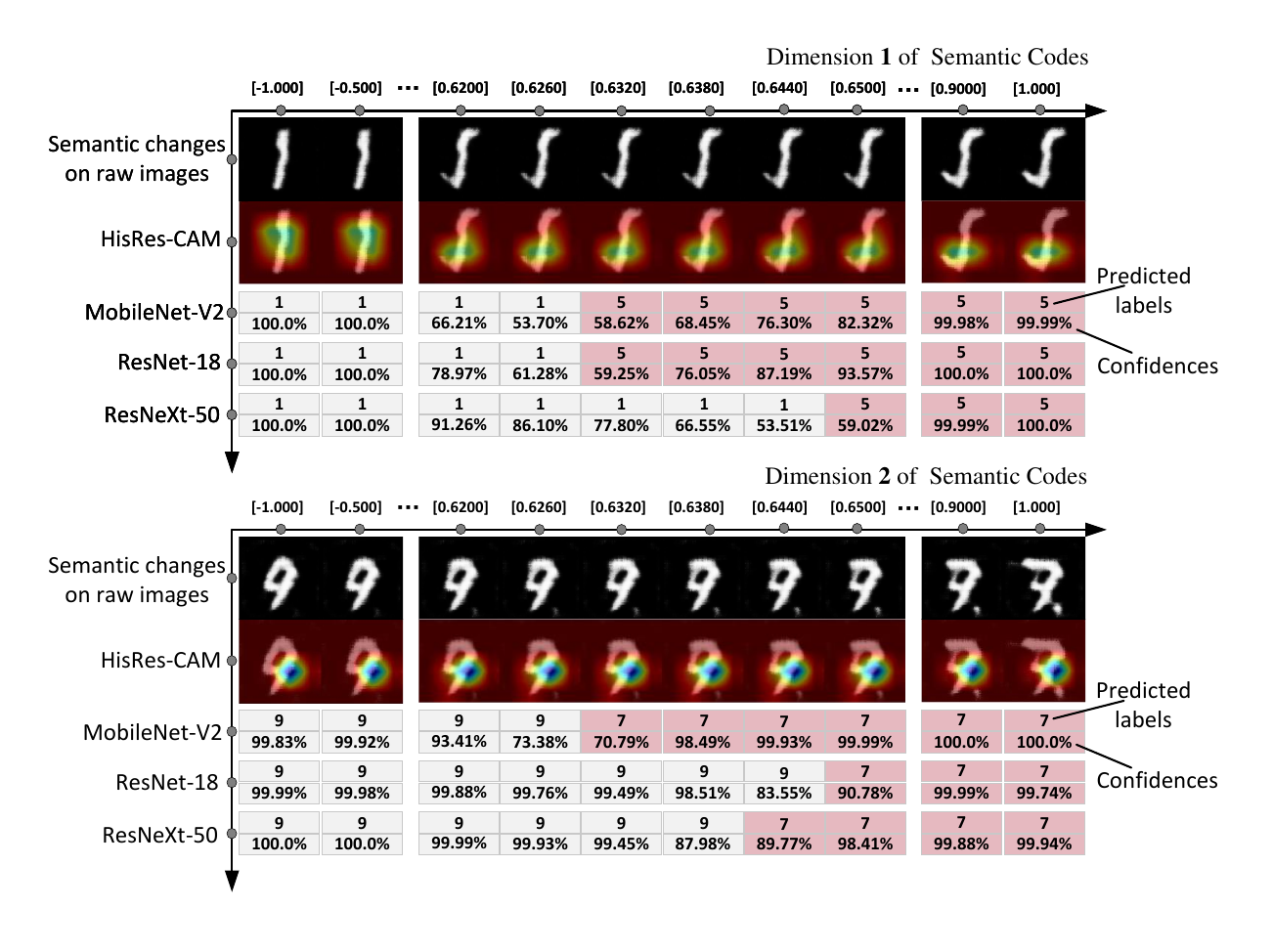}
    \vspace{-1.0cm}
    \caption{Continuous transitions created along a specific dimension of $z_3$. Red marked out incorrect classifications under well-trained models. Meanwhile, we also gave predicted labels and confidence scores under different models for each group of semantic variants.} 
    \label{fig-inter}
\end{figure*}\begin{table*}[!ht]
\centering
\setlength{\tabcolsep}{3.5pt}
\renewcommand{\arraystretch}{1.05}
\caption{Results of ASR-DM and ISR under various attacks, and two defenses including Adversarial Training and Network Distillation were employed to defend MobileNet-V2. For Adversarial Training, PGD adversary with $\Delta=3.0$, step size of $0.3$, and iterations of $12$ was employed to retrain and obtain the defended MobileNet-V2. For Network Distillation, we employed a novel parameter-free network that was identical to MobileNet-V2 and selected the distillation temperature to be $100$. \textbf{The bold and grayscale grids marked the optimal values under the various methods}.}
\vspace{-0.2cm}
\small
\begin{tabular}{c|l|c|c|c|c|l|c|c|c|c}
\hline
\hline
\specialrule{1pt}{0pt}{1pt}
& \multirow{3}{*}{\textbf{Distace$_{\ell_2}$}} & \multicolumn{4}{c|}{\textbf{MNIST}} & \multirow{3}{*}{\textbf{Distace$_{\ell_2}$}}&\multicolumn{4}{c}{\textbf{DEFECT}}                                                           
\\  \cline{3-6}\cline{8-11}
\textbf{Competitors}& 
& \multicolumn{2}{c|}{\textbf{Adversarial Training}} 
& \multicolumn{2}{c|}{\textbf{Network Distillation}} 
&
& \multicolumn{2}{c|}{\textbf{Adversarial Training}} 
& \multicolumn{2}{c}{\textbf{Network Distillation}}
\\ \cline{3-6}\cline{8-11}
& 
&\multicolumn{1}{c|}{\textbf{ASR-DM}}   
&\multicolumn{1}{c|}{\textbf{ISR}}
&\multicolumn{1}{c|}{\textbf{ASR-DM}}   
&\multicolumn{1}{c|}{\textbf{ISR}} 
&
&\multicolumn{1}{c|}{\textbf{ASR-DM}}   
&\multicolumn{1}{c|}{\textbf{ISR}}
&\multicolumn{1}{c|}{\textbf{ASR-DM}}   
&\multicolumn{1}{c}{\textbf{ISR}}
\\
\hline
\multirow{2}{*}{PGD$_{\ell_2}$}             
& ($\Delta = 3.0$)              
& 0.42 \%         
& 95.56 \%  
& 1.48 \%
& 97.04 \%
& ($\Delta = 3.0$)          
& 23.61 \% 
& 76.39 \%
& 49.84 \%
& 50.16 \%
\\

&  ($\Delta = 15.86$)               
&  12.03 \%          
&  87.97 \% 
& 58.65 \%
& 41.36 \%
& ($\Delta = 4.49$)            
& 20.10 \%  
& 79.90 \%
& 56.36 \%
& 43.64 \%
\\\hline
\multirow{2}{*}{APGD$_{\ell_2}$}             
& ($\Delta = 3.0$)              
& 0.85 \%
& 99.15 \%        
& 2.77 \% 
& 97.23 \%
& ($\Delta = 3.0$)            
& 19.25 \%  
& 80.75 \% 
& 59.05 \%
& 40.95 \%
\\

&  ($\Delta = 15.86$)               
& 52.34 \%           
& 47.66 \% 
& 92.98 \%
& 7.02 \%
& ($\Delta = 4.49$)           
& 16.69 \% 
& 83.31 \%
& 73.45 \%
& 26.55 \%
\\\hline

\multirow{2}{*}{Square$_{\ell_2}$}             
& ($\Delta = 3.0$)              
&  0.21 \%         
&  99.79 \% 
&  2.76 \%
& 97.24 \%
& ($\Delta = 3.0$)         
& 30.27 \%  
& 69.73 \%
& 52.36 \%
& 47.64 \%
\\

&  ($\Delta = 15.86$)               
&  40.17 \%           
&  59.83 \% 
&  61.31 \%
&  38.69 \%
& ($\Delta = 4.49$)         
& 39.55 \% 
& 60.65 \%
& 57.48 \%
& 42.52 \%
\\\hline
\multirow{2}{*}{AutoAttack$_{\ell_2}$}             
& ($\Delta = 3.0$)              
&  0.00 \%          
&  100.0 \% 
&  1.26 \%
& 98.74 \%
&  ($\Delta = 3.0$)          
& 18.99 \%  
& 81.01 \%
& 57.69 \%
& 42.31 \%
\\

&  ($\Delta = 15.86$)               
&  50.10 \%          
&  49.90 \%
& 88.63 \%
& 11.37 \%
&  ($\Delta = 4.49$)          
& 17.15 \%
& 82.85 \%
& 73.79 \%
& 26.21 \%
\\\hline

PGDRS$_{\ell_2}$             
&  ($\Delta = 15.86$)               
& 4.85 \%           
& 95.15 \% 
& 16.57 \%
& 83.43 \%
& ($\Delta = 4.49$)            
&  15.57 \% 
& 41.70 \%
& 46.89 \%
& 10.38 \%
\\\hline
DeepFool$_{\ell_2}$             
&  ($\Delta = 0.99$)               
&  0.21 \%         
&  98.31 \%
&  0.42 \%
&  98.10 \%
&  ($\Delta = 0.15$)         
&  21.48 \%  
&  78.52 \%
&  19.18 \%
&  80.82 \%
\\\hline
CW$_{\ell_2}$             
&  ($\Delta = 1.12$)               
&  23.20 \%          
&  76.80 \% 
&  26.57 \%
&  73.43 \%
& ($\Delta = 2.13$)           
& 18.61 \% 
& 81.39 \%
& 27.54 \%
& 72.46 \%
\\\hline
\multirow{2}{*}{MAELS}             
& ($\Delta <  15.86$)              
&  27.00 \%       
&  61.61 \%
&  25.94 \%
&  64.67 \%
& ($\Delta < 4.49$)            
&  11.07 \%
&  67.91 \%
&  20.33 \%
&  50.74 \%
\\

&  ($\Delta = 15.86$)               
&  \cellcolor{gray!25}{\textbf{87.34 \%}}       
&  \cellcolor{gray!25}{\textbf{1.27 \%}}
&  86.50 \%
&  \cellcolor{gray!25}{\textbf{2.11 \%}}
& ($\Delta = 4.49$)           
& \cellcolor{gray!25}{\textbf{74.75 \%}}  
& \cellcolor{gray!25}{\textbf{1.23 \%}}
& 70.46\%
& \cellcolor{gray!25}{\textbf{0.61 \%}}
\\ \hline
\hline
\specialrule{1pt}{0pt}{1pt}

\end{tabular}
\label{tab2}
\end{table*}\begin{figure*}[!ht]
    \centering  
    \includegraphics[width=1.0\textwidth]{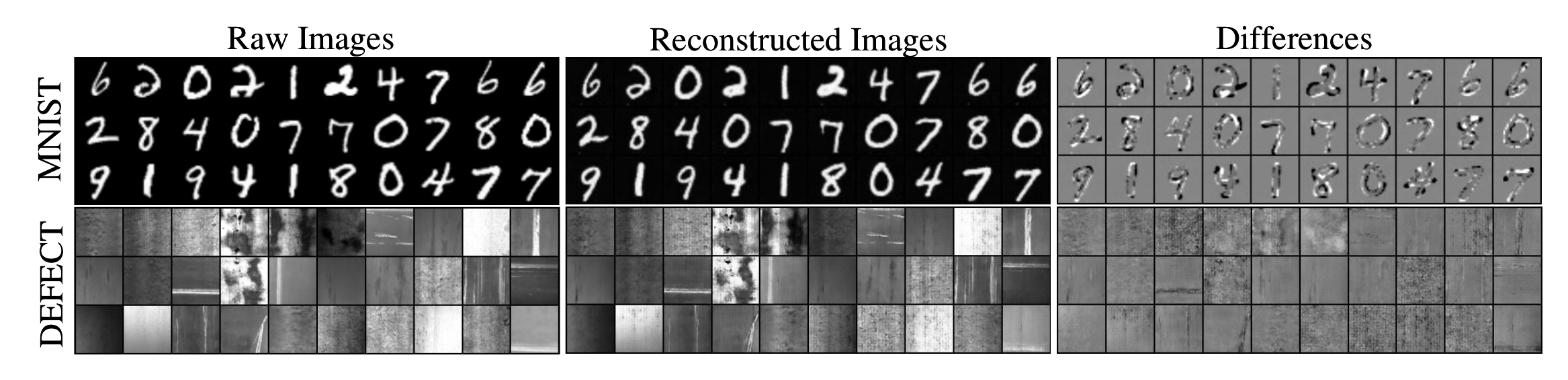}    
    \vspace{-0.8cm}
    \caption{Reconstruction Quality of reconstructed images relative raw images. Raw Images, Reconstructed Images, and Differences represent raw images, reconstructed images, and the differences between raw images and reconstructed images respectively.} 
    \label{fig-recon}
\end{figure*}
\subsection{Interpretable network vulnerabilities}
While existing pixel-wise attacks revealed vulnerabilities of DNNs, interpreting the vulnerabilities was a persistent challenge, due to the fact that these attacks could hardly be attributed to human perceptible modifications. Some structural perturbation works \cite{tao2018attacks, casper2022robust} had shown that attribute-level adversaries could reveal understandable weaknesses in the victim model. As mentioned in previous sections, our method manipulated the latent representations to create feature-level transformations and showed good potentials to explore perceptible interpretability. 

Fig. \ref{fig-inter} illustrated the continuous transitions we created along a specific dimension of $z_3$. We utilized HiResCAM heatmaps \cite{draelos2020use} to visualize the areas in the pixel space that had a greater impact on the model's decision-making process when interpreting the generated examples. Additionally, we presented the predicted labels and corresponding confidence scores for each generated example under various networks, with any incorrect predictions highlighted in red.

Take the digit $1$ as an example, under continuous control of $z_3$ in dimension \textbf{1}, its semantics in the bottom left corner gradually changed, as illustrated by a series of raw images. This subtle alteration finally made MobileNet-V2 misclassify it as digit $5$. Notably, as this attribute gradually changed from the leftmost raw image to the rightmost one, the confidence of successfully predicting it as the digit $1$ decreased from 100\% to 53.70\%, while the confidence of deeming it as the digit $5$ soared from 58.62\% to 99.99\%. A similar pattern could also be found in another example of the digit $9$. The model's decision boundary evidently exhibited susceptibility to these variations in semantics. Furthermore, this set of continuous samples would yield analogous outcomes when evaluating with the other two models, i.e., ResNet-18, and ResNeXt-50, albeit with variations in the magnitudes of semantic changes. These visualizations elucidated the superior perceptible interpretability of the MAELS. 
\subsection{Breaking Known Defenses}
The MAELS demonstrated a weaker dependence on specific victim models, accompanied by a higher transferability, as shown in Section \ref{4.2}, raising concerns on the effectiveness of some defense techniques. To assess the potential of the MAELS to bypass certain defenses, we conducted experiments on two recognized defense strategies: adversarial training, which involved training the network with the MAELS to enhance its resilience, and network distillation which aimed to mitigate the impact of bounded noise on raw images. We selected MobileNet-V2 for protection and tested it against the MAELS and seven other attackers. The evaluation indicators in this section included ASR for the defended MobileNet-V2 (ASR-DM), differing from the ASR in Tab. \ref{tab-asr} for the undefended MobileNet-V2, and the interception success rate (ISR), referring to the success rate of intercepting different attacks after applying security protection. All results were provided in Tab. \ref{tab2}, where higher ASR-DM and lower ISR values indicated a more aggressive approach. 

As shown in Tab. \ref{tab2}, compared to the ASR before applying any defenses (\textit{see} Tab. \ref{tab-asr}), ASR-DM experienced different levels of reduction, reflecting the effectiveness of the two defense methods. When adversarial training was applied, the MAELS retained ASR of 87.34\% and 74.75\% while exhibiting only ISR of 1.27\% and 1.23\% on MNIST and DEFECT, respectively. In contrast, ASR-DM of both MNIST and DEFECT on adversarial training decreased significantly. Similarly, when network distillation was employed, the MAELS had success rates of 86.50\% and 70.46\%. Although the ASR-DM of the MAELS under the defense of network distillation was not the highest among all methods, the ISRs for the MAELS under the two defense methods were the lowest and only 2.11\% and 0.61\% on MNIST and DEFECT, respectively, indicating that the MAELS successfully bypassed adversarial training and network distillation.

Since our method began with reconstructing images based on the raw inputs to create the final AEs, the raw images could be reconstructed flawlessly using the generative structures, achieving identical reconstruction became harder with increasing image resolutions. However, our strategy of replacing raw images with reconstructed ones for attacking purposes did not undermine the credibility of our experimental analysis. This was due to the significantly minimized differences before and after reconstruction, as demonstrated in Fig. \ref{fig-recon}. 
\section{Conclusion and Discussion}
In this paper, we presented a highly implementable method for generating AEs with legitimate semantics, called MAELS. Based on generative models within representation learning, we designed SSTGM, a semantic-transformation model, to manipulate the semantic attributes associated with image themes, generating a series of smooth and semantically coherent variations on a low-dimensional manifold. We focused on modifying only the relevant details of the image themes while preserving other non-thematic details along the semantic-oriented manifold. Extensive experiments conducted on two datasets demonstrated that the MAELS obtained both high visual quality and attack transferability to unknown models. Furthermore, the seamless transition from raw images to AEs exposed the vulnerabilities that led to decision failures in victim models, providing valuable insights for model interpretability and assessment of its robustness. Significantly, these AEs with legitimate semantics successfully bypassed two previously effective defenses, including adversarial training and network distillation.  

In future work, we will develop more sophisticated and refined semantic manipulation techniques for launching legitimate semantic-aware attacks, e.g., generating natural and legitimate AEs in the fields of face recognition and automatic driving. Meanwhile, we plan to generate hybrid AEs by integrating modifications for semantic features into point-based attacks, addressing the conflict between geometric distance and imperceptibility in point-based attacks.

\end{document}